\documentclass[10pt,twocolumn,letterpaper]{article}

\usepackage{cvpr}
\usepackage{times}
\usepackage{epsfig}
\usepackage{graphicx}
\usepackage{amsmath}
\usepackage{amssymb}

\usepackage{algorithm}

\usepackage{algpseudocode}

\usepackage[breaklinks=true,bookmarks=false]{hyperref}
\usepackage{caption} 
\cvprfinalcopy 

\def\cvprPaperID{****} 

\begin{document}

\title{MassFace: an efficient implementation using triplet loss for face recognition}

\author{Yule Li\\
{\tt\small yuzhile\_study@163.com}
}

\maketitle

\begin{abstract}
In this paper we present an efficient implementation using triplet loss for face recognition. We conduct the practical experiment to analyze the factors that influence the training of triplet loss. All models are trained on CASIA-Webface dataset and tested on LFW. We analyze the experiment results and give some insights to help others balance the factors when they apply triplet loss to their own problem especially for face recognition task. Code has been released in https://github.com/yule-li/MassFace.

\end{abstract}

\section{Introduction}

\begin{figure*}[t]
    \centering
    \includegraphics[width = 0.92\linewidth]{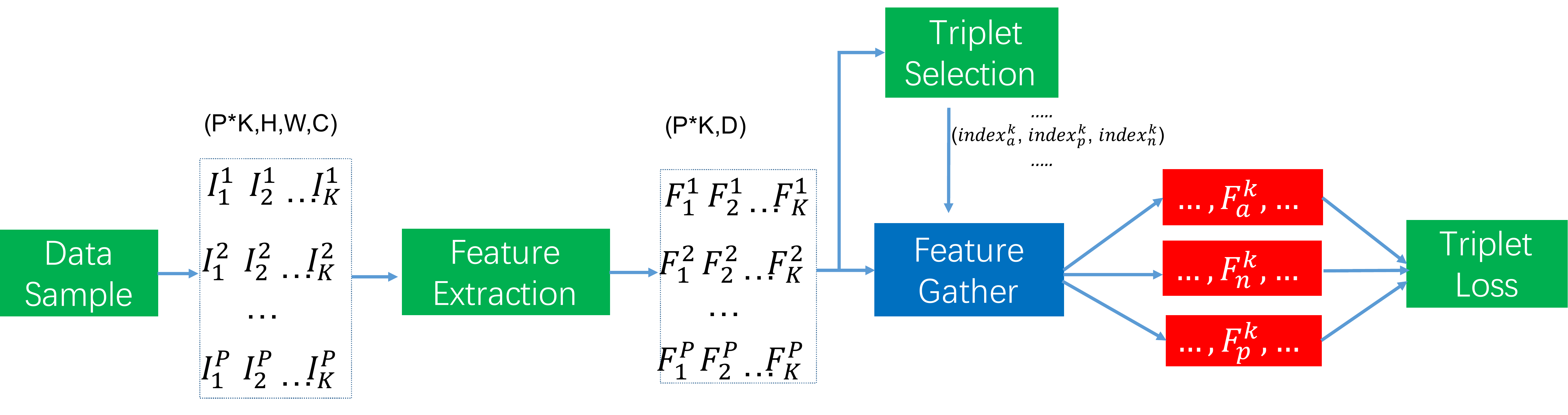}
    \caption{\small
      Framework using triplet loss to train face recognition model: total $P*K$ images are sampled for $P$ persons with $K$ images each person. The sampled images are mapped into feature vectors through deep convolutional network. The indexs of  triplet pairs are computed by a hard example mining process based on the feature vectors and the responding triplet feature pairs can be gathered. Finally, the features of triplet pairs are inputed into triplet loss to train the CNN.
    }
    \label{fig:method_show}
\end{figure*}

Face recognition has achieved significant improvement due to the power of deep representation through convolutional neural network. Convolutional neural network(CNN) based method first encodes the image which contains face into deep presentation and then apply the loss function to train the CNN such that the distance of feature vectors of the same persons is smaller than that of the different persons. Amost all of these loss functions can be divided into two categeries: 1) softmax classification based loss and its variants such as Sphere face~\cite{liu2017sphereface}, Arcface~\cite{arcface} and Cosface~\cite{wang2018cosface}; 2) metric learning based loss such as contrastive loss~\cite{sun2014deep} and Triplet loss~\cite{schroff2015facenet}. The previous one recently draw more attentions and has achieved great progress with the development of angular softmax loss and larger margin softmax loss to enhance the discriminative power of softmax loss. However, the last one is hard to train and heavily depends on people's experiences of hard example mining due to high computional complexity such as $O(N^3)$ of triplet loss for a dataset with $N$ samples.

In this work, we present an efficient implentation of triplet loss on face recognition task and conduct serveral experiments to analyze the factors that influence the training of triplet loss. The overview of our implementation can be seen in figure~\ref{fig:method_show}. Unlike the softmax loss, data sample of triplet loss need ensure that the valid triplet pair can be constructed as much as possible. In order to achieve this, we follow ~\cite{hermans2017defense} and the total $P*K$ images are sampled for $P$ persons with $K$ images each. The sampled images are mapped into feature vector $F^p_k$ through deep convolutional network such as Resnet~\cite{he2016deep} or Mobilenet~\cite{howard2017mobilenets}. Then the triplet pairs are selected by a hard mining process based on the embedded feature vectors. For convenience of implementation, we first get the indexs of triplet pairs and then gather their responding features. Finally, the gathered features of triplet pairs are fed into triplet loss to train the CNN.

The contributions of our paper can be summarized as: 1) we present an efficient implementation of our proposed framework which provides variant choices for the component of our proposed framewrok such as different CNN feature extractor,  different triplet pair selection method as in figure~\ref{fig:method_show} and also support multi-gpus to accelerate training process. 2) we practically  analyze a series of factors that influence training of triplet loss by experiments.

\section{Related Work}
\textbf{FaceNet: A Unified Embedding for Face Recognition and Clustering}~\cite{schroff2015facenet}. FaceNet uses traiplet loss to train the CNN model and mines semi-hard examples to train the triplet loss. It utilized on a dataset with 8M identities and trained on a cluster of cpu for thousands of hours. FaceNet achieves remarkable result on LFW~\cite{LFWTech} while it's not practical because it relys on such large dataset and need a large mount of time to  train

\textbf{In Defense of the Triplet Loss for Person Re-Identification}~\cite{hermans2017defense}. This work applys triplet loss on person re-identification problem. It constructs image batch efficiently by sampling $P$ persons with $K$ images each. It also proposed batch hard and batch all hard example mining  strategies. This work achieves the start-of-the-art in person re-identification.

\section{Framework}
The overview of our proposed framework can be viewed as figrue~\ref{fig:method_show}. The framework include the following module: data sample, feature embedding, triplet selection and triplet loss. We will describe them in details as follow.

\textbf{Data Sample}. Following ~\cite{hermans2017defense}, a batch of input images consists of $P$ persons and each person includes $K$ images. So in each iteration, we sample total $P*K$ images. Using such data sampling method, it's convenient to select valid triplet pairs and mine hard examples.

\textbf{Feature Extraction}. We use MobileFacenets~\cite{chen2018mobilefacenets} to extract the feature $F$ of the input image $x$ as a deep representation. We also fix the feature $||F_i||=1$ by L2-normalization. It only has about 6.0MB parameters and can be infered very fast as well.

\textbf{Triplet Loss}
Through the feature extraction powered by CNN, the input image $x$ can be mapped into a feature vector with $d$ dimension, and the map function is denoted as $f(x)$. The goal of the triplet loss is to make sure that the feature vector $f^a_i(x)$ of image $x^a_i$(called $anchor$) is close to $f^p_i(x)$ of image $x^p_i$(called $positive$) which has the same identity as image $x^a_i$ while $f^a_i(x)$ is far away from $f^n_i(x)$ of the image $x^n_i(x)$(called $negative$) that has the different identity as $x^a_i$. We can formulate this loss as:
\begin{equation}
\|f^a_i(x) -f^p_i(x)\|^2_2 + \alpha < \|f^a_i(x) -f^n_i(x)\|^2_2
\end{equation}
where $\alpha$ is the margin to avoid the collapse of $f_i(x)$. And the loss that will be optimized can become $L$
\begin{equation}
L=max(0,\|f^a_i(x) -f^p_i(x)\|^2_2 + \alpha - \|f^a_i(x) -f^n_i(x)\|^2_2)
\label{opt_loss}
\end{equation}

\textbf{Triplet Selection}
Triplet selection aims to choice the valid triplet $(i,j,k)$ which is used as input of triplet loss. The valid triplet means that $i$, $j$ have the identity and $i$, $k$ have different identity. As see in \textbf{Data sample}, the input is composed of $P$ persons with $K$ images each and total $B=P*K$ images. In order to obtain the all possible valid triplets, we iterate each image $x^k_a$ as anchor in person $k$ and any other image $x^k_p$ in person $k$ can be positive. Thus the negetive can be from all images  of other persons except k.  We summary this as algothrim ~\ref{all_triplets}. So there are about $O(P*K*P*K)$ valid triplet pairs but not every triplet pair can contribute to the triplet loss. As we see in formula \ref{opt_loss}, only the triplet pair that satisfies $\|f^a_i(x) -f^p_i(x)\|^2_2 + \alpha - \|f^a_i(x) -f^n_i(x)\|^2_2 > 0$ has loss value. We develop algorithm to mine such triplet pairs. We also develop different kinds of strategies to mine the 'hard' examples based on algorithm ~\ref{all_triplets}, and our experiment shows that these hard mining strategies can achieve better performanc for triplet loss. We summary all these mining strategies as follow.
\begin{itemize}
\item \textbf{Batch All}. As we can see in formula ~\ref{opt_loss}, only the triplet pair $(i,j,k)$ that satisify $\|f_i(x) -f_j(x)\|^2_2 + \alpha > \|f_i(x) -f_k(x)\|^2_2$ has the loss value. So we choice all of these triplet pairs as "hard" examples. We just modify  algorithm~\ref{all_triplets} a little to obtain the Batch All algorithm \ref{batch_all}.
\item \textbf{Batch Random} If there are many negatives for some $anchor$ and $positive$, we randomly select a negative. That can be destribed as  algorithm \ref{batch_random}.
\item \textbf{Batch Min Min} There may be many negatives for some $(anchor, positive)$ and we select the $negative$ that has the least distance with $anchor$. There may be also many $positives$ for some $anchor$, and we continue to select the $positive$ in which the responding $negative$ has least distance with $anchor$. That can be shown as algorithm \ref{batch_min_min}.
\item \textbf{Batch Min Max} There may be many $negatives$ for some $(anchor, positive)$ and we select the $negative$ that has the least distance with $anchor$. There may be also many positives for some $anchor$, and we just select the $positive $in which the responding $negative$ has the biggest distance with $anchor$. That can be shown as algorithm \ref{batch_min_max}.
\item \textbf{Batch Hardest} There may be many valid triplet pairs for some person, and we select only one pair in which $negative$ has the least distance with $anchor$. This can be seen as algorithm \ref{batch_hardest}.
\end{itemize}

\begin{algorithm}
\caption{Select all possible valid triplets}
\label{all_triplets}
\begin{algorithmic}[1]
\State {Random choice $B$ input images for $P$ persons with $K$ images each}
\State {Initilize list $T$ to hold all selected triplet pairs}
\For {Each person $p \in [1,P]$}
	\For {Each image $i$ in person $p$ as anchor}
		\For {Each image $j$ in person $p$ and $j!=i$ as positive }
			\For {Each image $k \in [1,B]$ and $k$ is not in person $p$}
				\State {T.append((i,j,k))}
			\EndFor
		\EndFor
	\EndFor
\EndFor

\end{algorithmic}
\end{algorithm}

\begin{algorithm}
\caption{Batch All}
\label{batch_all}
\begin{algorithmic}[1]
\State {Random choice $B$ input images for $P$ persons with $K$ images each}
\State {Forwarding $B$ input images to obtain the feature pool $F$}
\State {Compute distance matrix $M_{B\times B}$ between $F$}
\State {Initilize list $T$ to hold all selected triplet pairs}

\For {Each person $p \in [1,P]$}
	\For {Each image $i$ in person $p$ as anchor}
		\For {Each image $j$ in person $p$ and $j!=i$  as positive }
			\For {Each image $k \in [1,B]$ and $k$ is not in person $p$ as negative}
				\If {$M(i,j)+ \alpha > M(i,k)$}
					\State {T.append((i,j,k))}
				\EndIf
			\EndFor
		\EndFor
	\EndFor
\EndFor
\end{algorithmic}
\end{algorithm}



\begin{algorithm}
\caption{Batch Random}
\label{batch_random}
\begin{algorithmic}[1]
\State {Random choice $B$ input images for $P$ persons with $K$ images each}
\State {Forwarding $B$ input images to obtain the feature pool $F$}
\State {Compute distance matrix $M_{B\times B}$ between $F$}
\State {Initilize list $T$ to hold all selected triplet pairs}

\For {Each person $p \in [1,P]$}
	\For {Each image $i$ in person $j$ as anchor}
		\For {Each image $j$ in person $p$ and $j!=i$ as positive }
			\State {Initilize list t}
			\For {Each image $k \in [1,B]$ and $k$ is not in person $p$ as negative}
				\If {$M(i,j)+ \alpha > M(i,k)$}
					\State {t.append((i,j,k)}
				\EndIf
			\EndFor
			\State {Random choice $k$ in t and T.append($k$)}
		\EndFor
		
	\EndFor
\EndFor
\end{algorithmic}
\end{algorithm}

\begin{algorithm}
\caption{Batch Min Min}
\label{batch_min_min}
\begin{algorithmic}[1]
\State {Random choice $B$ input images for $P$ persons with $K$ images each}
\State {Forwarding $B$ input images to obtain the feature pool $F$}
\State {Compute distance matrix $M_{B\times B}$ between $F$}
\State {Initilize list $T$ to hold all selected triplet pairs}

\For {Each person $p \in [1,P]$}
	\For {Each image $i$ in person $p$ as anchor}
		\State {Initilize list t}
		\For {Each image $j$ in person $p$ and $j!=i$ as positive }
			\For {Each image $k \in [1,B]$ and $k$ is not in person $p$ as negative}
				\If {$M(i,j)+ \alpha > M(i,k)$}
					\State {t.append((i,j,k))}
				\EndIf
			\EndFor
		\EndFor
		\State {$k\_min\_min = \arg\min\limits_{(i,j,k)}{\{M(i,k)|(i,j,k) \in t\}}$}
		\State {T.append(k\_min\_min)}
	\EndFor
\EndFor
\end{algorithmic}
\end{algorithm}

\begin{algorithm}
\caption{Batch Min Max}
\label{batch_min_max}
\begin{algorithmic}[1]
\State {Random choice $B$ input images for $P$ persons with $K$ images each}
\State {Forwarding $B$ input images to obtain the feature pool $F$}
\State {Compute distance matrix $M_{B\times B}$ between $F$}
\State {Initilize list $T$ to hold all selected triplet pairs}

\For {Each person $p \in [1,P]$}
	\For {Each image $i$ in person $p$ as anchor}
		\State {Initilize list t1}
		\For {Each image $j$ in person $p$ and $j!=i$ as positive }
			\State {Initilize list t2}
			\For {Each image $k \in [1,B]$ and $k$ is not person $p$ as negative}
				\If {$M(i,j)+ \alpha > M(i,k)$}
					\State {t2.append((i,j,k))}
				\EndIf
			\EndFor
			\State {$k\_min = \arg\min\limits_{(i,j,k)}{\{M(i,k)|(i,j,k) \in t2\}}$}
			\State {t1.append($k\_min$)}
		\EndFor
		\State {$k\_min\_max = \arg\max\limits_{(i,j,k)}{\{M(i,k)|(i,j,k) \in t1\}}$}
		\State {T.append($k\_min\_max$)}
	\EndFor
\EndFor
\end{algorithmic}
\end{algorithm}

\begin{algorithm}
\caption{Batch Hardest}
\label{batch_hardest}
\begin{algorithmic}[1]
\State {Random choice $B$ input images for $P$ persons with $K$ images each}
\State {Forwarding $B$ input images to obtain the feature pool $F$}
\State {Compute distance matrix $M_{B\times B}$ between $F$}
\State {Initilize list $T$ to hold all selected triplet pairs}

\For {Each person $p \in [1,P]$}
	\State {Initilize list t}
	\For {Each image $i$ in person $p$ as anchor}
		\
		\For {Each image $j$ in person $p$ and $j!=i$ as positive }
			\For {Each image $k \in [1,B]$ and $k$ is not in  person $p$ as negative}
				\If {$M(i,j)+ \alpha > M(i,k)$}
					\State {t.append((i,j,k))}
				\EndIf
			\EndFor
		\EndFor
	\EndFor
	\State {$k\_min\_min = \arg\min\limits_{(i,j,k)}{\{M(i,k)|(i,j,k) \in t\}}$}
	\State {T.append(\_min\_min)}
\EndFor
\end{algorithmic}
\end{algorithm}

\textbf{Mining methods}. The triplet selection is based on a pool of the feature vectors $F$. There are serveral methods to obtain the pool of feature vectors with size B.
\begin{itemize}
\item \textbf{Online mining.} We obtain the pool of features by forwarding a batch of input images with size of B once a time.
\item \textbf{Offline mining.} We forword all images in dataset to get the pool of features and select triplet pairs. Then we train these triplet pairs by a sequence of iterations.
\item \textbf{Semi-online mining.} We generate the pool of features by forwording CNN model in serval iterations like $10$ times and then select triplet pairs. That can choice more triplet pairs while it doesn't consume too much time.
\end{itemize}



\section{Experiment}
We train all models on CASIA-Webface~\cite{yi2014learning} and test them on LFW. We first train the model with Softmax and CosFace respectively, as our pretrained model. Then all models based on triplet loss are optimized by ADAGRAD optimizer with learning rate $0.001$ and $\alpha$ is set as $0.2$. 

\textbf{ Results with different mining strategies}. We first pretrain CNN model with softmax classifer. Then we finetune the deep CNN model using triplet loss with different mining strategies. Every model is trained with $60k$ iterations and each iteration is optimized with batch size $210$. The results of triplet loss with different mining strategy can be seen as table~\ref{tab_strategy_exp}. All mining strategies can boost the performance of softmax classifer. The $BH\_min\_min$ strategy and $BH\_min\_max$ strategy improve performace more than BR and BA, while $B\_hardest$ is close to $BH\_min\_min$ and $BH\_min\_max$. The $BH\_min\_min$ strategy and $BH\_min\_max$ strategy is 'hard' strategy but not the 'hardest' strategy, which may make it easy to train for triplet loss.

\begin{table}[h]{}
\centering
\begin{tabular}{|c|c|}
\hline
strategy & acc(\%) \\\hline
softmax pretrain & 97.1 \\\hline
BH\_Min\_Min & 98.0 \\\hline
BH\_Min\_Max & 98.0 \\\hline
BH\_Hardest & 97.9 \\\hline
BH\_Random & 97.8 \\\hline
BH\_All & 97.5\\
\hline
\end{tabular}
\caption{Accuracy on LFW with different mining strategies.}\label{tab_strategy_exp}
\end{table}

\textbf{ Results with different initial models}. We also compare the models with different initial methods. We pretained two model with softmax classifer and CosFace respectively. Then we used these pretrained model as initial models and finetined it using triplet loss with $BH\_min\_max$ strategy which shows best performance in all strategies. The result can be viewed as table~\ref{tab_pretrain_exp}. The model with pretrain is more better than that without pretrain. The pretrained model gives a good start for triplet loss, which is essential for training of triplet loss. We also can see that pretrained model with CosFace is better than softmax because it has a better initialization.
\begin{table}[h]
\centering

\begin{tabular}{|c|c|c|}
\hline
pretrain& iters & acc(\%) \\\hline
softmax pretrain & 6w &97.1 \\\hline
cosface pretrain & 6w &98.3 \\\hline
with softmax & 6w &98.0 \\\hline
with cosface & 6w &98.6 \\\hline
without pretrain & 6w & 92.2 \\
\hline
\end{tabular}
\caption{Accuracy on LFW with different initial models.}\label{tab_pretrain_exp}
\end{table}

\textbf{Results with different ($P$,$K$) combinations}. In this experiment, we use the pretrained model trained by softmax and finetue it by triplet loss with the $BH\_min\_max$ strategy. The result of different combination can be seen as tabel~\ref{tab_pk_exp}. When we keep $B$($B=P*K$) as constant ($210$). From the table \ref{tab_pk_exp}, we can know that the larger the $P$ is, the better the performance is. For larger $P$, the $anchor$ can see more $negative$ from other persons, which may avoid the model trapped in local optimization.
\begin{table}[h]
\centering

\begin{tabular}{|c|c|c|}
\hline
P & K & acc(\%) \\\hline
42 & 5 & 98.0 \\\hline
30 & 7 & 98.0 \\\hline
14 & 15 & 97.7 \\\hline
10 & 21 & 97.5 \\
\hline
\end{tabular}
\caption{Accuracy on LFW with different $P$ and $K$ settings.}\label{tab_pk_exp}
\end{table}

\textbf{Results with different mining methods}. In this experiment, we compare the models trained in online and semi-online. All models are pretrained by Softmax and finetued with ${Batch\_min\_max}$ strategy. In simi-online, we forwarded the model by $10$ iterations with $210$ images each and then selected the triple pairs based on features of $10$ iterations. The results can be seen as table~\ref{tab_train_exp}. The sime-online training is better than online. This may be understood by that semi-online training increase the $P$ to improve the preformance as we see in tabel \ref{tab_pk_exp}. We demostrate this by add a experiment with multi-gpus. We increase batch size to $210*4$ by rising $P$ from $30$ to $30*4$ with 4 gpus. The accuracy of mutli-gpus is $98.3\%$, a similar accuracy with semi-online. We can use semi-online method to acheive the approaching performance of mutli-gpus when our computing resource is limited.
\begin{table}[h]
\centering
{}
\begin{tabular}{|c|c|}
\hline
train & acc(\%) \\\hline
softmax pretrain & 97.1 \\\hline
online & 98.0 \\\hline
semi-online & 98.2 \\\hline
online with multi-gpus & 98.3 \\
\hline
\end{tabular}
\caption{Accuracy on LFW with different mining methods.}\label{tab_train_exp}
\end{table}

\section{Conclusion}
We present an efficient implementation based on triplet loss for face recognition. We analyze the important factors that influence the performance of triplet loss by experiment. The results of experiment shows: 1) the pretrained model is very important for training CNN model with triplet loss; 2) hard example mining is essecial and we proposal two new mining methods: $BH\_min\_min$ and $BH\_min\_max$; 3) we can improve the performance by inreasing $P$ and a better way to do this is to train model with multi-gpus or in semi-online if the computing resource is limited.

\textbf{Acknowledgements.} We thank the supports from Key Lab of Intelligent Information Processing of Chinese Academy of Sciences and TAL-AILab.
{\small
\bibliographystyle{ieee}
\bibliography{egbib}
}

\end{document}